\documentclass[10pt,twocolumn,letterpaper]{article}

\usepackage[pagenumbers]{cvpr} 

\usepackage{comment}
\usepackage[dvipsnames]{xcolor}

\definecolor{cvprblue}{rgb}{0.21,0.49,0.74}
\usepackage[pagebackref,breaklinks,colorlinks,citecolor=cvprblue]{hyperref}

\def\authornote#1#2#3{{\textcolor{#2}{\textsl{\small#1:[#3]}}}}
\def\method{ZeroShape\xspace}

\newcommand{\zx}[1]{\authornote{Zixuan}{cyan}{#1}}

\def\paperID{4693} 
\def\confName{CVPR}
\def\confYear{2024}

\title{\method: Regression-based Zero-shot Shape Reconstruction}

\author{Zixuan Huang$^{1*}$ \quad Stefan Stojanov$^{2*}$ \quad Anh Thai$^{2}$ \quad Varun Jampani$^{3}$ \quad James M. Rehg$^{1}$\\
$^1$University of Illinois at Urbana-Champaign, \\$^2$Georgia Institute of Technology, $^3$Stability AI
}

\begin{document}
\maketitle
\def\thefootnote{*}\footnotetext{Both authors contributed equally to this work.}\def\thefootnote{\arabic{footnote}}
\begin{abstract}
We study the problem of single-image zero-shot 3D shape reconstruction. Recent works learn zero-shot shape reconstruction through generative modeling of 3D assets, but these models are computationally expensive at train and inference time. In contrast, the traditional approach to this problem is regression-based, where deterministic models are trained to directly regress the object shape. Such regression methods possess much higher computational efficiency than generative methods. This raises a natural question: is generative modeling necessary for high performance, or conversely, are regression-based approaches still competitive? To answer this, we design a strong regression-based model, called ZeroShape, based on the converging findings in this field and a novel insight. We also curate a large real-world evaluation benchmark, with objects from three different real-world 3D datasets. This evaluation benchmark is more diverse and an order of magnitude larger than what prior works use to quantitatively evaluate their models, aiming at reducing the evaluation variance in our field. We show that ZeroShape not only achieves superior performance over state-of-the-art methods, but also demonstrates significantly higher computational and data efficiency.\footnote{Project website at: \url{https://zixuanh.com/projects/zeroshape.html}}
\end{abstract}

\section{Introduction}
\label{sec:intro}

Inferring the properties of individual objects such as their category or 3D shape is a fundamental task in computer vision. The ultimate goal is to do this accurately for any object, generally referred to as zero-shot generalization.
For machine learning methods, this means high accuracy on data distributions that may be significantly different from the training set, such as images of novel types of objects like machine parts or images from uncommon visual contexts like underwater imagery.
An object representation capable of zero-shot generalization, therefore, needs to accurately capture the visual properties that are shared across all objects in the world---an extremely ambitious goal.



\begin{figure}[t]
\centering
    \includegraphics[width=0.95\linewidth]{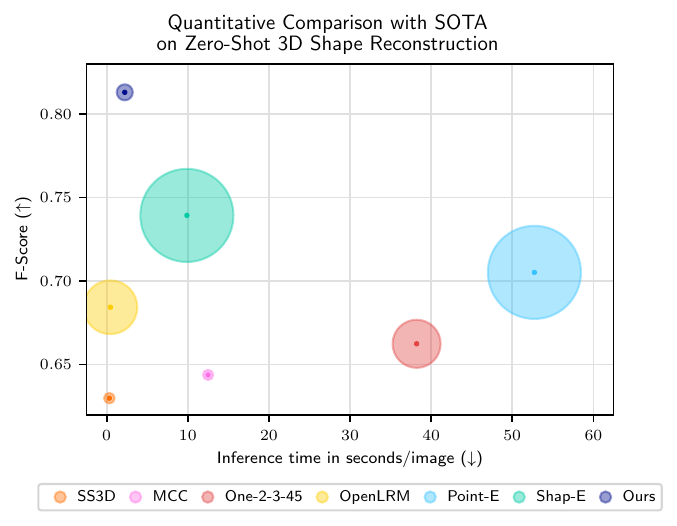}
    \vspace{-10pt}    
    \caption{We outperform SOTA methods for zero-shot 3D shape reconstruction, while having faster inference time and less training data. Circle size indicates the number of 3D assets used for training, with biggest being 3M\protect\footnotemark. F-Score with threshold 0.05 is averaged over Octroc3D~\cite{shrestha2022ocrctoc}, Pix3D~\cite{sun2018pix3d} and OmniObject3D~\cite{wu2023omniobject3d}.}
    \label{fig:bubble-teaser}
\end{figure}
\footnotetext{We use 3M as a reference value. Point-E~\cite{nichol2022point} and Shape-E~\cite{jun2023shap} state a dataset size of ``several million".}

\begin{figure*}[t]
\centering
	\includegraphics[width=\linewidth]{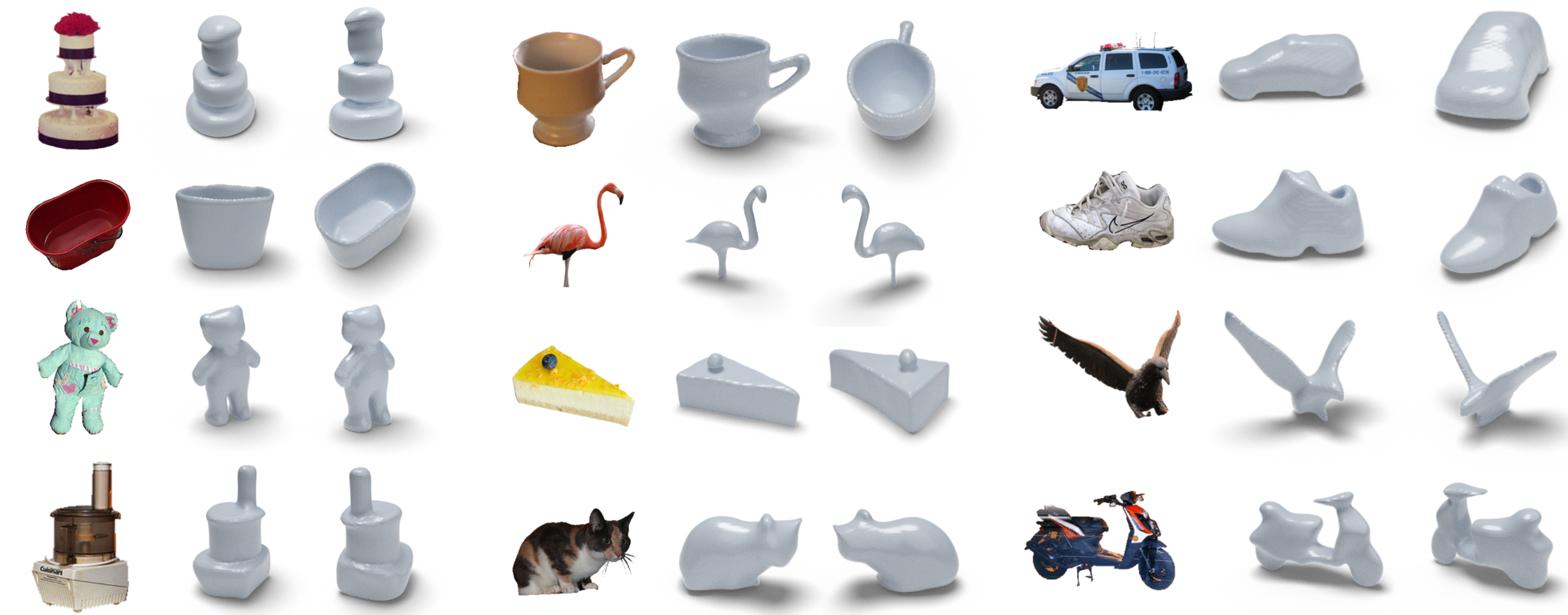}
	\caption{\textbf{ZeroShape reconstructions from in-the-wild images.} Our method produces detailed and accurate object reconstructions from single-view images on a diverse set of objects.}
	\label{fig:teaser}
\end{figure*}

Recent work in computer vision has taken the broader challenge of zero-shot generalization head-on, with impressive developments for 2D vision tasks like segmentation~\cite{kirillov2023segany,qi2023cropformer}, visual question answering~\cite{alayrac2022flamingo, li2023blip}, image generation~\cite{rombach2021highresolution, midjourney, saharia2022photorealistic}, and in training general vision representations that can be easily adapted for any vision task~\cite{radford2021clip,oquab2023dinov2}.
This progress has largely been enabled by increasing model size and scaling training dataset size to the order of tens to hundreds of millions of images.

These developments have inspired efforts which aim at zero-shot generalization for single image 3D object shape reconstruction~\cite{liu2023one, liu2023zero, nichol2022point, jun2023shap}.
This is a classical and famously ill-posed problem, with important applications like virtual object placement in scenes in AR and object manipulation in robotics. 
These works aim to learn a ``zero-shot 3D shape prior'' by relying on generative diffusion models for 3D point clouds~\cite{nichol2022point}, NeRFs~\cite{jun2023shap}, or for 2D images fine-tuned for novel-view synthesis~\cite{liu2023zero,liu2023one}, enabled by million-scale 3D data curation efforts such as Objaverse~\cite{deitke2022objaverse,deitke2023objaverse}. 
While these methods show impressive zero-shot generalization ability, it comes at a great compute cost due to large model parameter counts and the inference-time sampling required by diffusion models.
Using expensive generative modeling for zero-shot 3D shape from single images diverges from the approach of early deep learning-based works on this task~\cite{shin2018pixels,zhang2018learning,thai20213d,wu2023multiview,Xian2022gin}.
These works define the task as a 3D occupancy or signed distance regression problem and predict the shape of objects in a single forward pass.
This raises a natural question: is generative modeling necessary for high performance at learning zero-shot 3D shape prior, or conversely, can a regression-based approach still be competitive? 

In this work, we find that regression approaches are indeed competitive if designed carefully, and computationally more efficient by a large margin compared to the generative counterparts. We propose ZeroShape: a regression-based 3D shape reconstruction approach that achieves state-of-the-art zero-shot generalization, trained entirely on synthetic data, while requiring a fraction of the compute and data budget of prior work (see~\cref{fig:bubble-teaser}). 
We build our model upon key ingredients that facilitate generalization based on prior works: 
1) usage of intermediate geometric representation (e.g. depth)~\cite{marr2010vision,wu2017marrnet,zhang2018learning,thai20213d,Xian2022gin}, 
2) explicit reasoning with local features~\cite{xu2019disn,bechtold2021fostering,wu2023multiview}. 
Specifically, we decompose the reconstruction into estimating the shape of the visible portion of the object, and then predicting the complete 3D object shape based on this initial prediction.
The accurate estimation of the visible 3D surface is enabled by a joint modeling of camera intrinsics and depth,  which we find to be essential for high accuracy.





Another thrust of our work is a large benchmark for evaluating zero-shot reconstruction performance. 
The 3D vision community is working on developing a zero-shot 3D shape prior, but what is the correct way to evaluate our progress? 
Currently we lack a well-defined benchmark, which has lead to well-curated qualitative results and small scale quantitative results\footnote{On the order of hundreds of objects from tens of categories at best, to just a few dozen objects at worst.} on different datasets across different papers. 
This makes it difficult to track progress and identify directions for future research.
To resolve this and standardize evaluation, we develop a protocol based on data generated from existing datasets of 3D object assets.
Our benchmark includes thousands of common objects from hundreds of different categories and multiple data sources.
We consider real images paired with 3D meshes~\cite{sun2018pix3d, shrestha2022ocrctoc}, and also generate photorealistic renders of 3D object scans~\cite{wu2023omniobject3d}. Our large scale quantitative evaluation provides a rigorous perspective on the current state-of-the-art.

In summary, our contributions are:
\begin{itemize}
    \item ZeroShape: A regression-based zero-shot 3D shape reconstruction method with state-of-the-art performance at a fraction of the compute and data budget of prior work.
    \item A unified large-scale evaluation benchmark for zero-shot 3D shape reconstruction, generated by standardized processing and rendering of existing 3D datasets.
\end{itemize}

\section{Related Work}
\label{sec:related}
Estimating the 3D shape of an object from a single is a complex inverse problem: while the shape of the visible object can be estimated from shading, estimating the shape of the occluded portion requires prior knowledge about object geometry. 
This is one of the marvels of human perception and achieving it computationally is a major goal for our field. 
We review regression and generative methods for this task.

\noindent \textbf{Regression-based Methods.}
These works investigate different ways to represent 3D object shapes and the architectures to produce them from a single image, e.g., meshes~\cite{kanazawa2018learning, wang2018pixel2mesh, wen2019pixel2mesh++} or implicit representations like discrete~\cite{choy20163d, tatarchenko2017octree} or continuous~\cite{mescheder2019occupancy, peng2020convolutional} occupancy, signed distance fields~\cite{xu2019disn, thai20213d, irshad2022shapo}, point clouds~\cite{fan2017point, achlioptas2018learning}, or sets of parametric surfaces~\cite{groueix2018, yang2018foldingnet}. 
A major limitation of these works is the limited generalization beyond the categories of the training set. The improvements of decomposing the problem into predicting the depth and then estimating the complete shape~\cite{shin2018pixels,zhang2018learning,thai20213d,wu2017marrnet,Xian2022gin}, and representing 3D in a viewer centered rather than object centered reference frame~\cite{shin2018pixels,zhang2018learning,thai20213d} allowed for improved zero-shot generalization.
Most architectures follow an encoder/decoder design, where the encoder produces a feature map from which the decoder predicts the 3D shape. 
While early works produced a single feature vector for the entire image, it was later identified that using local features from a 2D feature map improved the detail of the predicted shapes~\cite{wang2018pixel2mesh, xu2019disn} and improved generalization to unseen categories~\cite{Xian2022gin, bechtold2021fostering}. 
This culminated with the current state-of-the art regression method, MCC~\cite{wu2023multiview}, which takes an RGB-D image as input, and uses a transformer-based encoder-decoder setup to produce a ``shell occupancy'' prediction~\footnote{Traditionally occupancy is formulated as predicting whether a point in 3D is inside/outside a watertight mesh, whereas MCC predicts whether it is within an $\varepsilon$ wide shell representing the surface of the object.}.
Our approach incorporates all these prior findings for improved generalization, and builds upon them with a new module for estimating the visible shape of the object that estimates depth and camera intrinsics, which is processed with a cross attention-based decoder to produce an occupancy prediction.

\noindent \textbf{3D Generative Methods} This category of methods does zero-shot 3D shape reconstruction using a learned 3D generative prior, where the 3D generation is conditioned on one or few input images. Given image or text conditioning, early work~\cite{wu2018learning} used GANs to generate voxels, whereas more recent works use diffusion models to generate point clouds~\cite{nichol2022point}, or function parameters for implicit 3D representations~\cite{jun2023shap}. 
Another related type of generative framing is conditional view synthesis. Works in this direction fine-tune 2D generative models~\cite{liu2023zero}, or train them from scratch~\cite{watson2022novel,zhou2022sparsefusion}, to synthesize novel views conditioned on single images and viewing angles. This results in an implicit 3D prior, from which a 3D shape can then be extracted by fitting a 3D neural representation to the synthesized images, or predicting its parameters~\cite{liu2023one}. 

\noindent \textbf{3D from 2D Generative Models} There have been efforts to use the real-world 2D image prior from text-to-2D generative models~\cite{rombach2021highresolution, saharia2022photorealistic, ramesh2022hierarchical, midjourney} to reconstruct 3D shape from a single image. 
This category of works~\cite{melaskyriazi2023realfusion,deng2022nerdi, tang2023make} often uses techniques such as the SDS loss~\cite{poole2022dreamfusion, wang2023score} and generates 3D assets from images by optimizing for each object separately. 
The prolonged optimization time prevents these works from being evaluated at scale or applied in many real-world applications. 
Orthogonal to the optimization-based approaches, we focus on learning a 3D shape prior that generalizes across instance. We do not perform any per-instance optimization at test time.

\section{Method}
\label{sec:method}

Our goal is to achieve state of the art zero-shot performance for estimating the complete 3D shape of an object from a single image. Formally, given an object-centric single-view RGB image $I \in \mathbb{R}^{h \times w \times 3}$, we regress a function that takes $I$ as input and predicts the shape. We represent shape using an implicit occupancy representation, where we model the shape surface as the isosurface of occupancy function $f(\boldsymbol{x}|I; \theta)$. Here, $\boldsymbol{x} \in \mathbb{R}^{3}$ denotes the query point's coordinates---when the query point lies within the surface $f(\boldsymbol{x}|I; \theta) = 1$, otherwise $f(\boldsymbol{x}|I; \theta) = 0$.

\subsection{Architecture}
\begin{figure*}[t]
\centering
	\includegraphics[width=\linewidth]{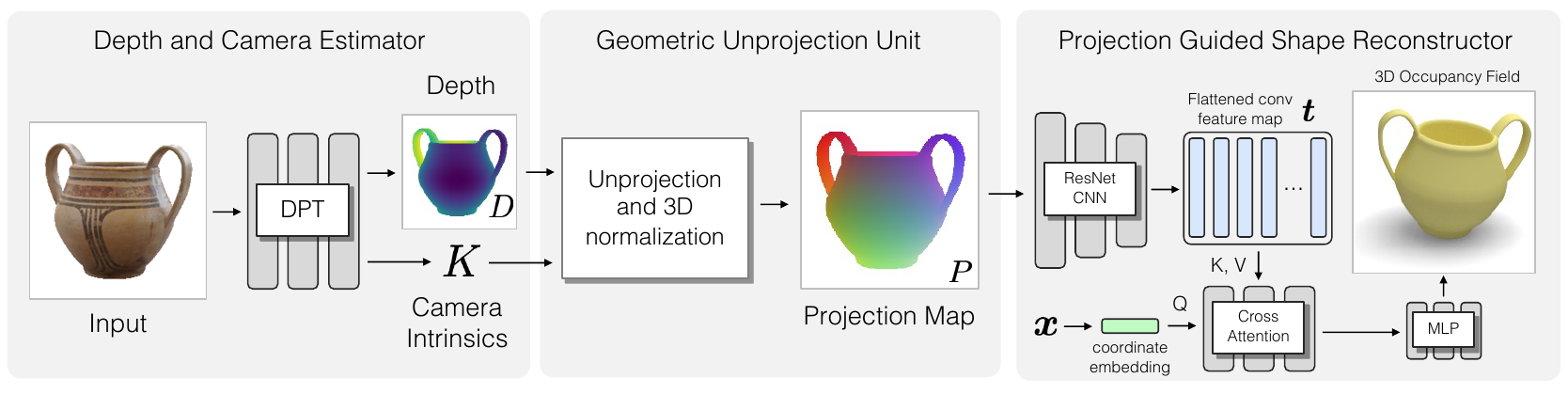}
	\caption{\textbf{Overview of our model.} Our consists of three modules: a depth and camera estimator, a geometric unprojection unit and a projection-guided shape reconstructor. The depth and camera estimator predicts the depth and camera intrinsics from the input image with a DPT backbone. The geometric unprojection unit converts the depth and intrinsics estimation into a normalized 3D visible surface, which is parameterized by a three-channel projection map. The shape reconstructor finally reconstructs the full occupancy field by fetching localized information from projection map through cross attention.}
	\label{fig:arch}
\end{figure*}
We now present our architecture (see~\cref{fig:arch}) for shape reconstruction. Our architecture is based on two established practices from prior works in this field: 1) usage of intermediate geometric representation~\cite{marr2010vision,wu2017marrnet,zhang2018learning,thai20213d,Xian2022gin} and 2) explicit reasoning with spatial feature maps~\cite{xu2019disn,bechtold2021fostering,wu2023multiview}. Specifically, our model consists of three submodules: a depth and camera estimator, a geometric unprojection unit and a projection-guided shape reconstructor. 

\noindent \textbf{Depth and camera estimator.}
We propose to estimate the 3D visible object surface as an intermediate representation.
To infer the full shape of an object, one must understand the visible surface---not only because the visible surface is often a large part of the full surface, but also because an accurate visible surface facilitates geometric reasoning of the full object reconstruction. 
This is because cues for reconstruction that allow for generalization, such as symmetry, curvature, and repetition, can be more effectively detected and leveraged in the 3D space.
For example, if an object is symmetric, then accurately inferring the 3D symmetry planes from a partial 3D surface is much easier than from 2D RGB or relative depth. 

\begin{figure}[h]
\centering
    \includegraphics[width=0.95\linewidth]{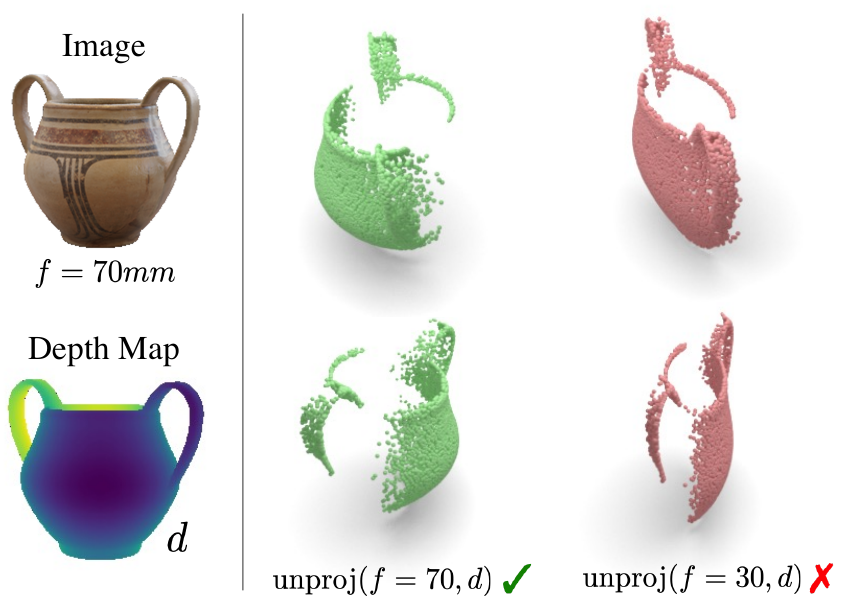} 
    \caption{\textbf{Effect of Intrinsics.} Unprojecting an accurate depth map into a 3D surface surface with erroneous intrinsics leads to skewed shape with wrong 3D aspect ratio.}
    \label{fig:intrinsics_prediction}
\end{figure}

Inspired by this, we estimate 3D visible surface as our intermediate representation instead of the commonly used depth maps~\cite{wu2017marrnet,zhang2018learning,thai20213d}. MCC~\cite{wu2023multiview} also uses visible surface to estimate the full shape, but they assume the visible surface to be given as input. When inferring on in-the-wild images, they use fixed intrinsics to unproject depth maps into the 3D surface. Erroneous intrinsics lead to skewed 3D visible surfaces (see~\cref{fig:intrinsics_prediction}), resulting in inaccurate 3D cues for the complete object shape. Therefore, we propose to jointly estimate depth and intrinsics before predicting the full shape. Note that learning to estimate depth and intrinsics can be fully supervised with synthetic data.
Specifically, our depth and camera estimator estimates the depth map of the object $D \in \mathbb{R}^{h \times w}$ and the camera intrinsics $K \in \mathbb{R}^{3 \times 3}$ from the image $I$. We used a shared DPT~\cite{Ranftl2021} backbone for the depth and camera estimator, and use two different shallow heads to predict $D$ from the local tokens and $K$ from the global token.

\noindent \textbf{Geometric unprojection unit.}
Given the intrinsics $K$ and the depth map $D$, the geometric unprojection unit unprojects them into a projection map $P \in \mathbb{R}^{h \times w \times 3}$. The projection map encodes the visible surface of the object, where each pixel value $P_{ij}$ represents the coordinate of the unprojected 3D point at the pixel location $(i, j)$. Formally, the geometric unprojection can be written as
\begin{equation}
    P_{ij} = D_{ij} K^{-1} {[i, j, 1]}^T.
    \label{eq: unproj}
\end{equation}
We use a view-centric coordinate system, because prior works show that view-centric learning is beneficial to generalization~\cite{tatarchenko2019single,thai20213d}. Therefore the camera coordinate frame is the “world” coordinate frame for shape reconstruction, which means that only the camera intrinsics matrix is required to unproject pixels to 3D. Note that unprojection is fully differentiable w.r.t. $D$ and $K$, so we can easily use it as a module in an end-to-end learning-based model. Additionally, the projection maps are foreground-segmented, and the represented visible surface is normalized in the 3D space to be zero-mean and unit-scale before being fed into the next module.

\noindent \textbf{Projection-guided shape reconstructor.}
Using the estimated projection map $P$, our projection-guided shape reconstructor recovers the full object shape. Specifically, the projection-guided shape reconstructor first uses a ResNet~\cite{he2016deep} encoder to encode and reshape the projection map $P$ into a set of $d$-dimensional vectors, $\boldsymbol{t} \in \mathbb{R}^{k \times d}$. Each of the $k$ vectors encodes a localized patch in the projection map. To facilitate an explicit spatial reasoning, we use the cross-attention-based approach proposed in MCC~\cite{wu2023multiview}. We linearly map every query point $\boldsymbol{x} \in \mathbb{R}^3$ to the same dimension of feature vectors, $d$. Then we use two cross attention layers to fetch relevant patch encodings from $\boldsymbol{t}$ and fuse them with each query separately. Finally, the reconstructor predicts the occupancy value of each $\boldsymbol{x}$ using the fused feature vector via an MLP.

\subsection{Loss}
We use a two-stage training paradigm for our model, where we first pretrain the depth and camera estimator and then fine-tune the whole model with 3D supervision. For depth and camera pretraining, we use a depth loss $\mathcal{L}_{depth}$ and projection-based intrinsics loss $\mathcal{L}_{proj}$. For the depth loss, we use the SSIMAE loss from~\cite{ranftl2020towards}. Note that the SSIMAE loss is scale-invariant, meaning that the depth estimator trained using this loss will be metrically correct up to an unknown scale factor. Therefore, directly regressing the absolute intrinsics is suboptimal due to the uncertainty in the absolute scale. 
Instead, we observe that the only factor that impacts shape reconstruction is whether the visible surface recovered using~\cref{eq: unproj} is accurate. Therefore, we directly minimize the MSE loss between the predicted projection map $P$ and the ground truth projection map $P^*$, and backpropagate to the camera and depth estimators.


Once the depth and camera estimator are learned, we jointly train the whole model using the 3D occupancy loss $\mathcal{L}_{occ}$, which is a standard binary cross entropy between the predicted occupancy values $f(\boldsymbol{x}|I; \theta)$ and ground truth in the viewer-centric coordinate frame. 

\subsection{Implementation Details}
We train our model with the Adam~\cite{kingma2014adam} optimizer. During depth and camera pretraining, we use a learning rate of $3 \times 10^{-5}$, a batch size of 44, a weight decay of 0.05 and momentum parameters of (0.9, 0.95). We train our model for 15 epochs and initialize the depth estimator with the Omnidata~\cite{eftekhar2021omnidata} weights. During the joint training stage, we use a learning rate of $3 \times 10^{-5}$ for the projection-guided shape reconstructor, and a learning rate of $10^{-5}$ for the pretrained depth and camera estimator (geometric unprojection unit does not have learnable parameters). We use a batch size of 28, a weight decay of 0.05 and momentum parameters of (0.9, 0.95). At every iteration, we randomly sample 4096 points to compute the occupancy loss. We train our model on $4\times$ NVIDIA GeForce RTX 2080 Ti, which takes $\sim$2 days for pretraining and $\sim$3 days for joint training.

\section{Data Curation}

\subsection{Training Dataset} 
We use all the 55 categories of ShapeNetCore.v2~\cite{chang2015shapenet} for a total of about 52K meshes, as well as over 1000 categories from the Objaverse-LVIS~\cite{deitke2022objaverse} subset. This subset of Objaverse has been manually filtered by crowd workers to primarily include meshes of objects instead of other assets like scans of large scenes and buildings. After filtering Objaverse-LVIS to remove objects with minimal geometry (e.g. objects consisting of a single plane) this dataset has ~42K meshes. Pooling these two data sources gives us a total of over 90K 3D object meshes from over 1000 categories.

We use Blender~\cite{blender} to generate synthetic images from the 3D meshes, and to extract a variety of useful annotations: depth maps, camera intrinsics, and object and camera pose. Because the object distribution of ShapeNet is highly skewed (67\% of data is 7 categories), we generate 1 to 20 images per object, scaled inversely from the number of meshes in the category of the object, resulting in a total of 159K images. For Objaverse we generate 25 images per object resulting in 939K images. Our traning set consists of slightly less than 1.1M images.

We generate images with varying focal lengths, from 30mm to 70mm for a 35mm image sensor size equivalent. We generate diverse object-camera geometry: rather than the common approach of always pointing the camera at the middle of the object at a fixed distance, we vary the object camera distance and vary the LookAt point of the camera. This allows us to capture a wide range of variability in how 3D shape projects to 2D. We follow the convention to use center cropped and foreground segmented images for training and testing. We provide more details in the supplement.

\subsection{Evaluation Benchmark} 
We use three different real-world dataset evaluation: OmniObject3D~\cite{wu2023omniobject3d}, Ocrtoc3D~\cite{shrestha2022ocrctoc}, and Pix3D~\cite{sun2018pix3d}. Because our testing images images come from the real world, or are renders of real 3D object scans distinct from our training set, they are a good test set for zero-shot generalization.

\noindent \textbf{OmniObject3D.}
OmniObject3D is a large and diverse dataset of 3D scans and videos of objects from 216 categories, including household objects and products, food and toys. Because the foreground segmentations are noisy, we follow convention and render the 3D scans to generate test images~\cite{liu2023one, liu2023zero}. We improve the default material shader which generates glass-like surface appearance to appear more natural. We use Blender and HDR environment maps to generate realistic images with diverse lighting. We randomly sample camera viewpoint, distance and focal length.

\noindent \textbf{Ocrtoc3D.}
Ocrtoc3D is a real-world object dataset that contains object-centric videos and full 3D annotations from 15 coarse categories. Some coarse categories contain many subcategories (e.g. toy animals contain various species). For each video the mesh (3D scan) and the viewpoint information are provided. We clean up this dataset by manually removing outliers (e.g. empty meshes/wrong object scales) and use the full filtered dataset consisting of 749 unique image-object pairs.

\begin{figure*}[t]
\centering
	\includegraphics[width=\linewidth]{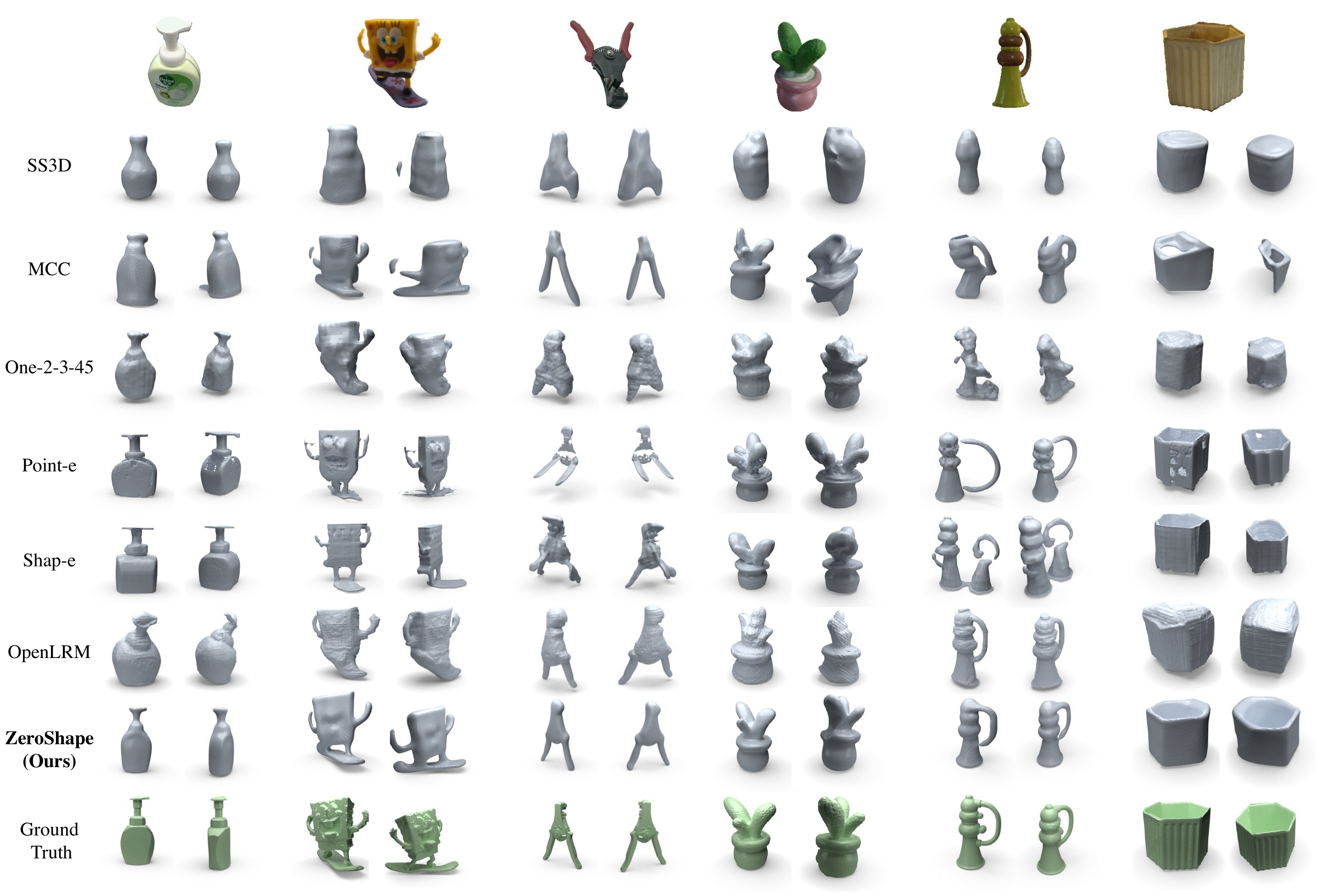}
	\caption{\textbf{Qualitative results.} We compare \method to other SOTA methods on our curated benchmark (first three columns are from Ocrtoc3D~\cite{shrestha2022ocrctoc}, last three are from OmniObject3D~\cite{wu2023omniobject3d}). Our reconstruction not only better aligns with the visible surfaces from images, but also recovers a faithful global structure of the reconstructed objects.}
	\label{fig:qualitative}
 \vspace{-10pt}
\end{figure*}

\noindent \textbf{Pix3D.}
Pix3D is a real-world object dataset that contains 3D annotations from 9 categories. For each image in this dataset, an object mask, a CAD model, and the input viewpoint information are provided. These 3D annotations come from manual alignment between shapes and images. We follow the split of~\cite{huang2022planes} and use 1181 images.

\noindent \textbf{Benchmark curation.} 
To create an easy-to-use benchmark, we convert the three heterogeneous datasets into a unified format. This includes aligning and converting the camera intrinsics and extrinsics, and object poses, to a standardized convention across the test datasets and our synthetic dataset. This is often a tedious obstacle in 3D vision research. We also organize images, masks and other metadata in a standardized manner. The release of our training data, data generating pipeline, and benchmark will benefit the community by providing a unified setup for large scale training on synthetic data and large scale testing on real data.
\section{Experiments}
\label{sec:experiments}
In this section we present our experiments, which include state-of-the-art comparisons and ablations. We first describe the baselines we implemented on our benchmark, and then show detailed quantitative and qualitative results.

\subsection{Metrics}
We evaluate the shape reconstruction models using Chamfer Distance (CD) and F-score as our quantitative metrics following~\cite{lin2020sdf,groueix2018,tatarchenko2019single,thai20213d,huang2022planes,huang2023shapeclipper}.

\noindent \textbf{Chamfer Distance.}
Chamfer Distance (CD) measures the alignment between two pointclouds. Following~\cite{huang2023shapeclipper}, CD is defined as an average of accuracy and completeness. Denoting pointclouds as $X$ and $Y$, CD is defined as:

\begin{equation}
    \small
    \text{CD}(X, Y) = \frac{1}{2|X|}\sum_{x \in X} \min_{y \in Y} \|x-y\|_2 + \frac{1}{2|Y|}\sum_{y \in Y} \min_{x \in X} \|x-y\|_2
\end{equation}

\noindent \textbf{F-score.} 
F-score measures the pointcloud alignment under a classification framing. By selecting a rejection threshold $d$, F-score@$d$ (FS@$d$) is the harmonic mean of precision@$d$ and recall@$d$. Specifically, precision@$d$ is the percentage of predicted points that lies close to GT point cloud within distance $d$. Similarly, recall@$d$ is the percentage of ground truth points that have neighboring predicted points within distance $d$. FS$@d$ can be intuitively interpreted as the percentage of surface that is correctly reconstructed under a certain threshold $d$ that defines correctness.

\noindent \textbf{Evaluation protocol.}
To compute CD and F-score, we grid-sample the implicit function and extract the isosurface via Marching Cubes~\cite{lorensen1987marching} for methods using implicit representation. Then we sample 10K points from the surfaces for the evaluation of CD and F-scores. Because most methods cannot predict view-centric shapes, we use brute-force search to align the coordinate frame of prediction with ground truth before calculating the metrics. This evaluation protocol ensures a fair comparison between methods with different shape representation and coordinate conventions.

\subsection{Baselines}
We consider five state-of-the-art baselines for shape reconstruction, SS3D~\cite{alwala2022pre}, MCC~\cite{wu2023multiview}, Point-E~\cite{nichol2022point}, Shap-E~\cite{jun2023shap}, One-2-3-45~\cite{liu2023one} and OpenLRM~\cite{hong2023lrm,openlrm}.

\noindent \textbf{SS3D} learns implicit shape reconstruction by first pretraining on ShapeNet GT, and then finetuning on real-world single-view images. The finetuning is performed in a category specific way, and then a single unified model is distilled from all category-specific experts. We compare our model with their final distilled model.

\noindent \textbf{MCC} learns shell occupancy reconstruction using multi-view estimated point clouds from CO3D~\cite{reizenstein2021co3d}. Their model assumes known depth and intrinsics during inference. To evaluate their model on RGB images, we use the DPT-estimated depth and fixed intrinsics as MCC's input following their pipeline.

\begin{table}[t]
\centering
\caption{\textbf{Quantitative comparison on OmniObject3D.} Our method performs favorably to other state-of-the-art methods. }
\begin{tabular}{lccc|c}
\hline
Methods & FS@1$\uparrow$ & FS@2$\uparrow$ & FS@5$\uparrow$ & CD$\downarrow$    \\ \hline
SS3D~\cite{alwala2022pre} & 0.1515 & 0.3482 & 0.6618 & 0.482 \\ 
MCC~\cite{wu2023multiview} & 0.1362 & 0.3215 & 0.6015 & 0.551 \\ 
One-2-3-45~\cite{liu2023one} & 0.1532 & 0.3585 & 0.6882 & 0.446 \\ 
OpenLRM~\cite{hong2023lrm} & \underline{0.1683} & \underline{0.3848} & \underline{0.7204} & \underline{0.407} \\
Point-E~\cite{nichol2022point} & 0.1505 & 0.3598 & 0.6932 & 0.448 \\
Shap-E~\cite{jun2023shap} & 0.1483 & 0.3650 & 0.7029 & 0.434 \\
ZeroShape (ours) & \textbf{0.2297} & \textbf{0.4927} & \textbf{0.8169} & \textbf{0.310} \\ \hline
\end{tabular}

\label{table:omni3d}
\end{table}

\noindent \textbf{Point-E} is a point cloud diffusion model that generates point clouds from text prompts or RGB images. They additionally train a separate model that converts point clouds into meshes. We compare our model with Point-E by combining their image-to-point and point-to-mesh models.

\noindent \textbf{Shap-E} is another diffusion model that learns conditioned shape generation from text or images. Different from Point-E, Shap-E uses a latent diffusion setup and can directly generate implicit shapes. The final mesh reconstruction are extracted with marching cubes.

\begin{table}[t]
\centering
\caption{\textbf{Quantitative comparison on Pix3D.} Our method performs favorably to other state-of-the-art methods. }
\begin{tabular}{lccc|c}
\hline
Methods & FS@1$\uparrow$ & FS@2$\uparrow$ & FS@5$\uparrow$ & CD$\downarrow$    \\ \hline
SS3D~\cite{alwala2022pre} & 0.1326 & 0.2998 & 0.6316 & 0.485 \\ 
MCC~\cite{wu2023multiview} & 0.1754 & 0.3386 & 0.6165 & 0.514 \\ 
One-2-3-45~\cite{liu2023one} & 0.1364 & 0.3137 & 0.6666 & 0.443 \\ 
OpenLRM~\cite{hong2023lrm} & 0.1458 & 0.3190 & 0.6440 & 0.492 \\ 
Point-E~\cite{nichol2022point} & 0.1779 & 0.3830 & 0.7255 & 0.403 \\
Shap-E~\cite{jun2023shap} & \textbf{0.2016} & \underline{0.4287} & \textbf{0.7833} & \textbf{0.340} \\
ZeroShape (ours) & \underline{0.1928} & \textbf{0.4290} & \underline{0.7759} & \underline{0.345} \\ \hline
\end{tabular}

\label{table:pix3d}
\end{table}

\noindent \textbf{One-2-3-45} learns implicit shape reconstruction by breaking it down into a generative view synthesis step and a multiview-to-3D reconstruction step. The view synthesis is achieved with Zero-123~\cite{liu2023zero}, a diffusion model that generates novel-view images conditioned on the original images and poses. Based on the synthesized multi-view images, a cost-volume-based module reconstructs the full 3D mesh of the object.

\noindent \textbf{LRM} is a concurrent work that learns to predict NeRF~\cite{mildenhall2020nerf} from single images using transformer-based architecture. Since the authors have not released the code, we use the code and weights from \textbf{OpenLRM}~\footnote{\url{https://github.com/3DTopia/OpenLRM}}. The mesh is extracted via Marching Cubes~\cite{lorensen1987marching} from the triplane NeRF.

\subsection{Comparison to SOTA Methods}
We compare our approach to other state-of-the-art methods on the benchmark we curated. We now present and analyze the quantitative results for each dataset.

\noindent \textbf{Results on OmniObject3D.}
We present our main quantitative comparison results on OmniObject3D, which covers a great variety of object types. The results are shown in~\cref{table:omni3d}. Comparing with other SOTA zero-shot 3D reconstruction methods, we see our approach achieves significantly better performance.

\noindent \textbf{Results on Ocrtoc3D.}
\begin{table}[t]
\centering
\caption{\textbf{Quantitative comparison on Ocrtoc3D.} Our method performs favorably to other state-of-the-art methods. }
\begin{tabular}{lccc|c}
\hline
Methods & FS@1$\uparrow$ & FS@2$\uparrow$ & FS@5$\uparrow$ & CD$\downarrow$    \\ \hline
SS3D~\cite{alwala2022pre} & 0.1271 & 0.2910 & 0.5963 & 0.543 \\
MCC~\cite{wu2023multiview} & \underline{0.1994} & \underline{0.4098} & 0.7135 & 0.411 \\
One-2-3-45~\cite{liu2023one} & 0.1323 & 0.3076 & 0.6325 & 0.492 \\
OpenLRM~\cite{hong2023lrm} & 0.1552 & 0.3481 & 0.6885 & 0.432 \\
Point-E~\cite{nichol2022point} & 0.1589 & 0.3591 & 0.6968 & 0.423 \\
Shap-E~\cite{jun2023shap} & 0.1725 & 0.3939 & \underline{0.7315} & \underline{0.395} \\
ZeroShape (ours) & \textbf{0.2410} & \textbf{0.5091} & \textbf{0.8459} & \textbf{0.286} \\ \hline
\end{tabular}

\label{table:ocrtoc3d}
\end{table}
We present additional quantitative comparison results on Ocrtoc3D. Ocrtoc is smaller than OmniObject, but still covers many object types, and the input images are real photos. The results are shown in~\cref{table:ocrtoc3d}. Similar to the results on OmniObject3D, our approach outperforms previous SOTA methods by a large margin.

\noindent \textbf{Results on Pix3D.}
We also present quantitative comparison results on Pix3D. Unlike OmniObject3D and Ocrtoc3D, the object variety of this evaluation dataset is much lower --- all objects are furniture and more than two third of the images are chairs and sofas. Therefore, the evaluation results are highly bias towards this specific class of objects. The results are shown in~\cref{table:pix3d}, and our method still achieves state-of-the-art performance. It is worth noting that Point-E and Shap-E also perform well on this dataset. We hypothesize this is might relate to the abundance of similar furniture categories in their training set. 

\subsection{Qualitative Results}
We show qualitative results of different methods in~\cref{fig:qualitative}. 
Generative approaches such as Point-E and Shap-E tend to have sharper surfaces and contain more details in their generation. However, many details are erroneous hallucination that do not accurately follow the input image, and the visible surfaces are often reconstructed incorrectly. 
Previous regression-based approaches such as MCC better follow the input cues in the input images, but the hallucination of the occluded surfaces is often inaccurate. 
We observe that One-2-3-45, OpenLRM and SS3D cannot always accurately capture details and concavities.
Comparing with prior arts, the reconstruction of ZeroShape not only faithfully capture the global shape structure, but also accurately follows the local geometry cues from the input image. More qualitative results are included in the supplement.


\begin{table}[t]
\centering
\caption{\textbf{Ablation study on OmniObject3D.} The design choices of our architecture are quantitatively justified: enforcing explicit geometric reasoning, and implementing it through unprojection with estimated depth and intrinsics is essential.}
\begin{tabular}{lccc|c}
\hline
Methods & FS@1$\uparrow$ & FS@2$\uparrow$ & FS@5$\uparrow$ & CD$\downarrow$    \\ \hline
Ours \textit{w/o geo} & 0.2110 & 0.4572 & 0.7797 & 0.347 \\ 
Ours \textit{w/o unproj} & 0.2135 & 0.4738 & 0.8053 & 0.323 \\ 
Ours \textit{w/o intr} & 0.2158 & 0.4742 & 0.8039 & 0.324 \\ 
Ours & \textbf{0.2297} & \textbf{0.4927} & \textbf{0.8169} & \textbf{0.310} \\ \hline
\end{tabular}
\label{table:ablation}
\vspace{-10pt}
\end{table}
\subsection{Ablation Study}
We analyze our method by ablating the design choices we made. We consider baselines by modifying different modules correspondingly. The results are shown in~\cref{table:ablation}.

\noindent \textbf{Explicit geometric reasoning.} 
We first consider the baseline without any geometric reasoning (Ours \textit{w/o geo}). We remove the projection unit together with the depth and camera pretraining losses. The number of parameters is controlled to be the same, and we train the model for the same number of total iterations. Comparing the first row to the last row, we see that enforcing explicit geometric reasoning in our model positively affects performance.

\noindent \textbf{Alternative intermediate representations.} 
Prior works~\cite{wu2017marrnet,wu2018learning,thai20213d} typically consider depth as the 2.5D intermediate representation. To compare this to our projection-based representation, we consider a baseline where the latent vectors directly come from the depth map instead of a 3D projection map. As shown in~\cref{table:ablation} (Ours \textit{w/o unproj}), depth leads to inferior performance to our intrinsic-guided projection map representation.

\begin{figure}[t]
\centering
    \includegraphics[width=\linewidth]{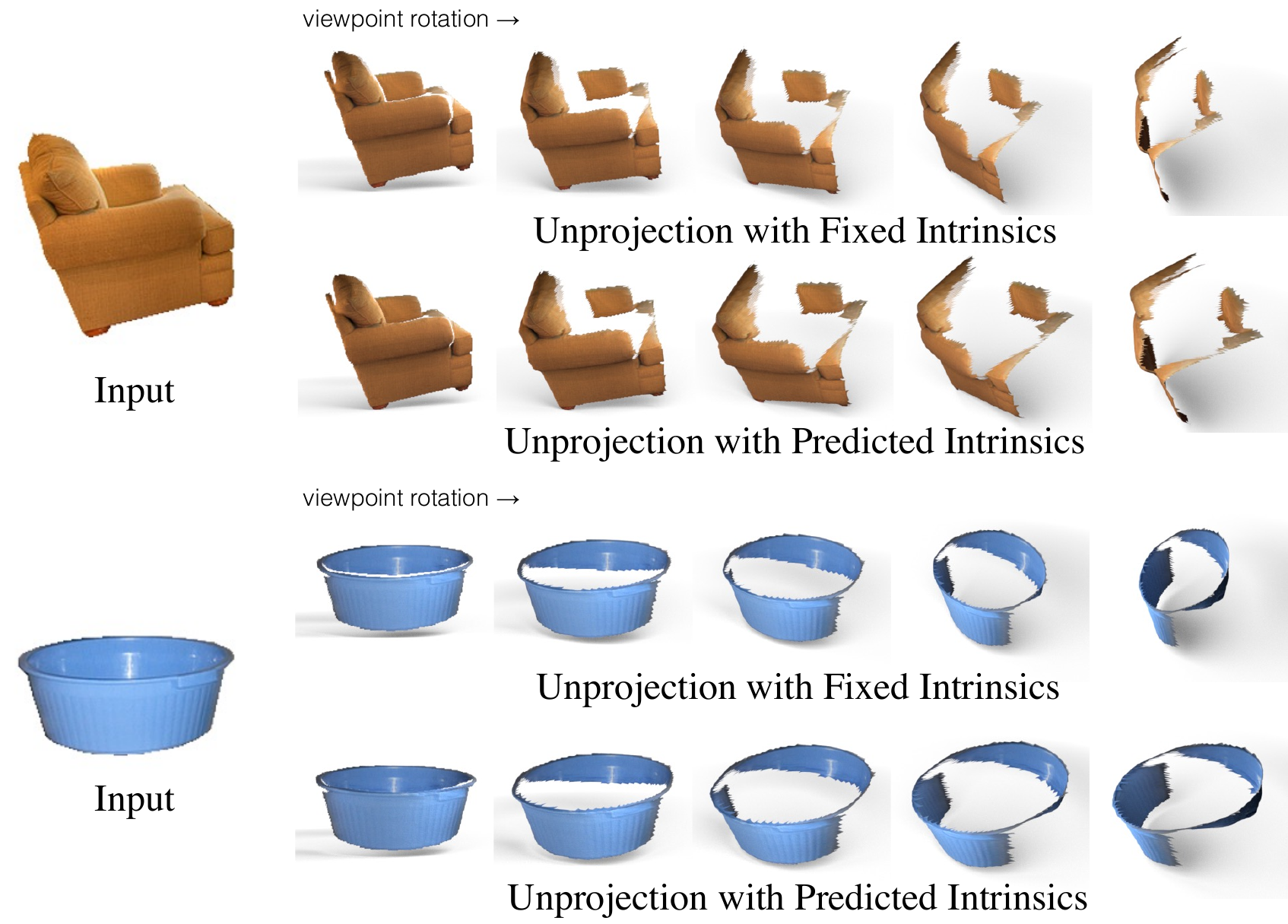} 
    \caption{\textbf{Benefits of intrinsics learning.} We show the reconstructed visible surfaces for two real image inputs. The visible surface is unprojected from estimated depths, with either fixed intrinsics or predicted intrinsics. Using fixed intrinsics cause unrealistic deformations in the 3D aspect ratio of the visible object surface (e.g. objects appear to be compressed).}
    \label{fig:intrinsics_qualitative}
\end{figure}

\noindent \textbf{Intrinsic-guided projection.}
We propose joint learning of intrinsics with depth to more accurately estimate the 3D shape of the visible object surface. To study the impact of this, we compare our full model with a baseline without intrinsics learning, where the unprojection to 3D is done via a fixed intrinsics during both training and testing. This baseline (Ours \textit{w/o intr}) leads to indifferent performance to using depth intermediate representation and is worse than our full model. We also show qualitative examples of the estimated surface using our pretrained intrinsics estimator in~\cref{fig:intrinsics_qualitative}. Compared with fixed intrinsics, unprojection with our estimated intrinisics leads to more accurate reconstruction of the visible surface.


\section{Limitations and Discussion}
\label{sec:limitation}

Due to computational resource limitations, we are not able to process and train our model on the full Objaverse dataset. Currently, the meshes from Objaverse we use only consist of 5\% of Objaverse and 0.4\% of Objaverse-XL objects. Based on the promising scaling properties of recent foundation models~\cite{dosovitskiy2020image,kaplan2020scaling,wei2022emergent}, we believe it will be valuable to explore the scaling properties of method. 

Another limitation of our work is that we have not considered the modeling of object texture. Predicting textures of unseen surfaces is highly ill-posed and can greatly benefit from a strong 2D prior. 
Given the recent success of 2D diffusion models~\cite{rombach2021highresolution} and their application in optimization-based 3D generation methods~\cite{poole2022dreamfusion,melaskyriazi2023realfusion,deng2022nerdi,lin2023magic3d,Chen_2023_ICCV,wang2023prolificdreamer}, we think it will be promising to initialize or regularize these methods with our shape prior, potentially boosting both the optimization efficiency and generation quality.

\section{Conclusion}
\label{sec:conclusion}

We present a strong regression-based model for zero-shot shape reconstruction.
The core of our model is an intermediate representation of the 3D visible surface which facilitates effective explicit 3D geometric reasoning.
We also curate a large real-world evaluation benchmark to test zero-shot shape reconstruction methods.
Our benchmark pools data from three different real-world 3D datasets and has an order of magnitude larger scale than the test sets used by prior work.
Tested on our benchmark, our model significantly outperforms other SOTA methods and achieves higher computational efficiency, despite being trained with much less 3D data.
We hope our effort is a meaningful step towards building zero-shot generalizable 3D reconstruction models.

\noindent \textbf{Acknowledgement.} This work was supported by NIH \\R01HD104624-01A1.

{
    \small
    \bibliographystyle{ieeenat_fullname}
    \bibliography{main}
}

\clearpage
\appendix

\begin{figure*}[h]
\centering
    \includegraphics[width=\linewidth]{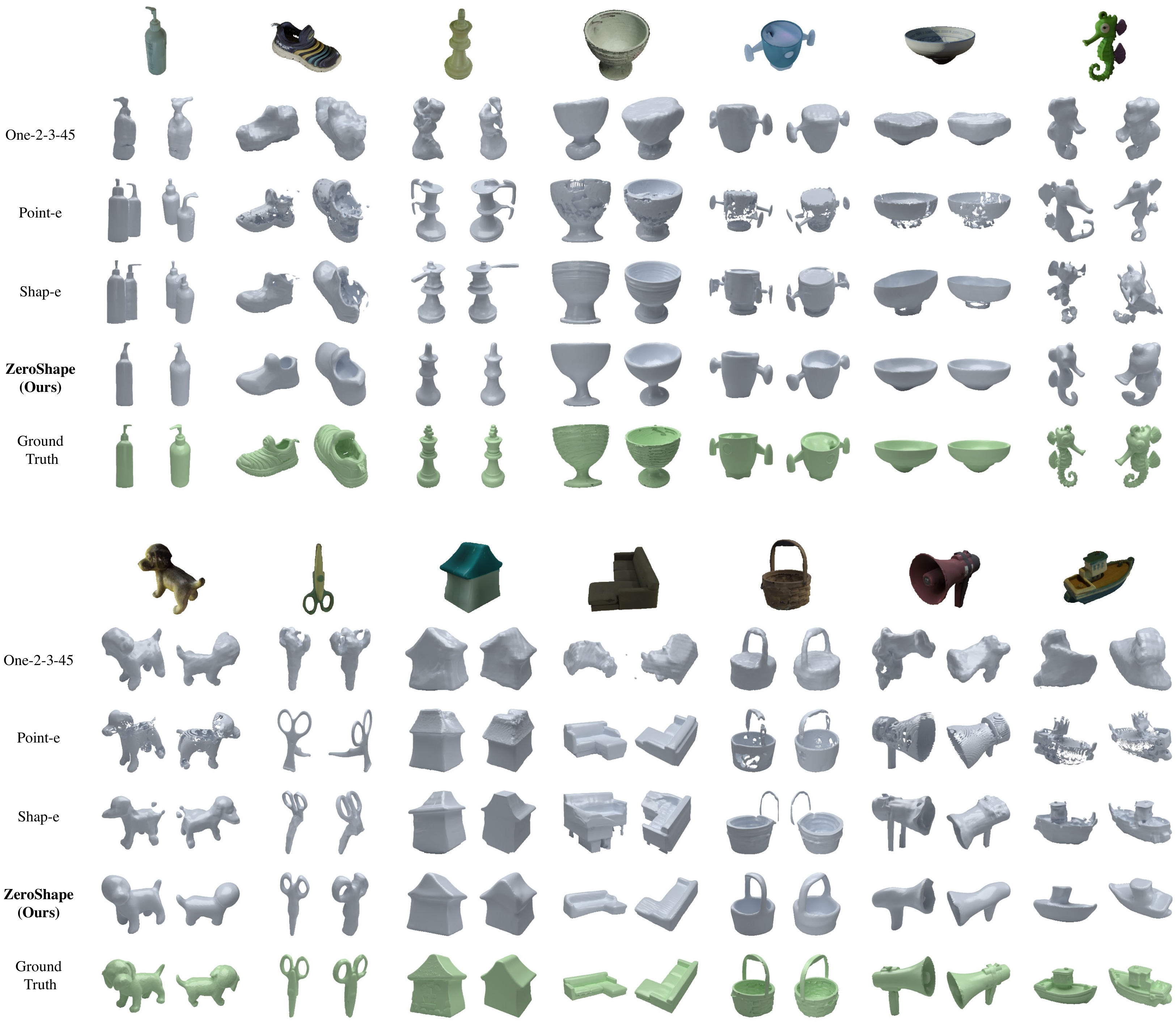} 
    \caption{Additional qualitative results and comparison on OmniObject3D.}
    \label{fig:quali-omni-supp}
\end{figure*}

\begin{figure*}[h]
\centering
    \includegraphics[width=\linewidth]{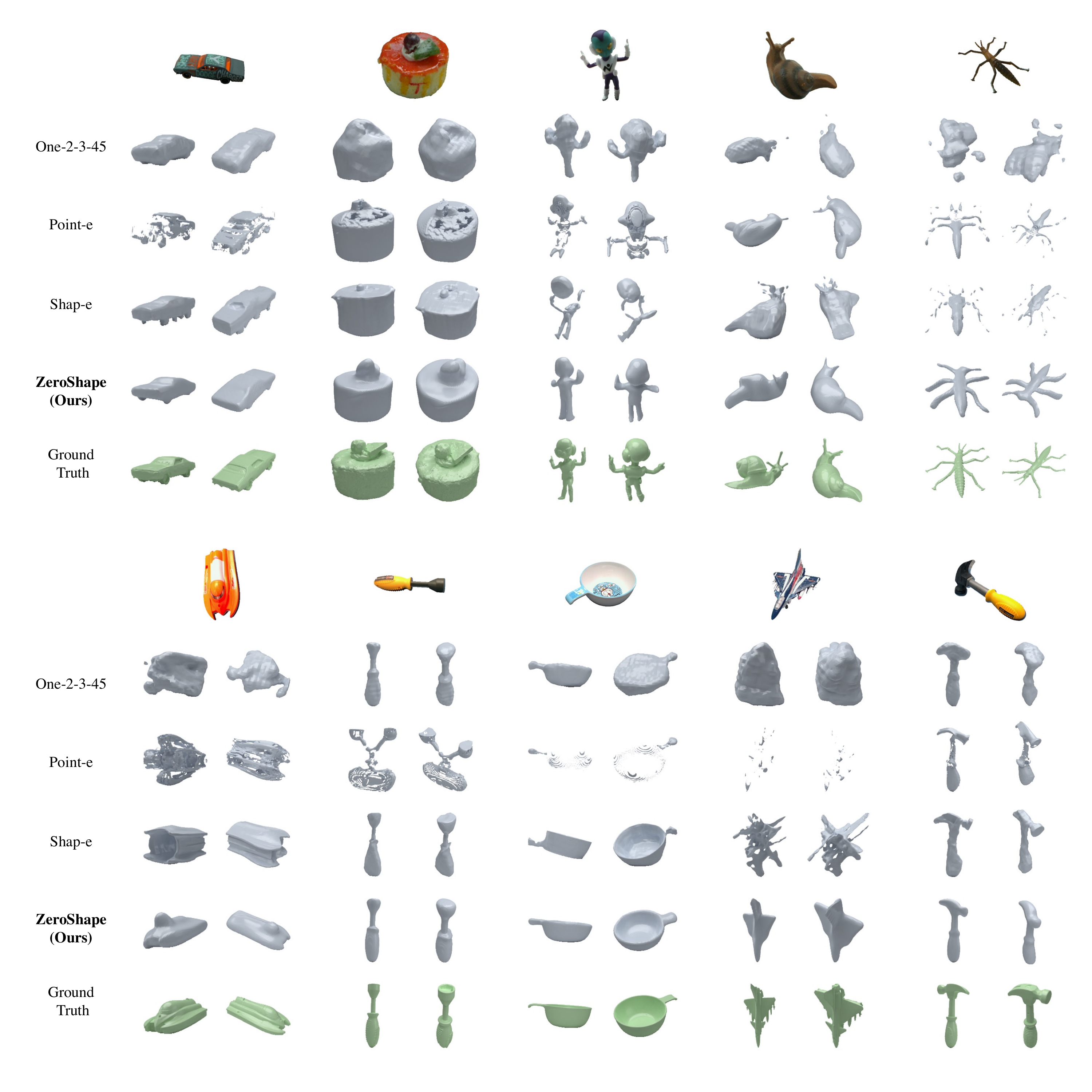} 
    \caption{Additional qualitative results and comparison on Ocrtoc3D.}
    \label{fig:quali-ocrtoc-supp}
\end{figure*}

\begin{figure*}[h]
\centering
    \includegraphics[width=\linewidth]{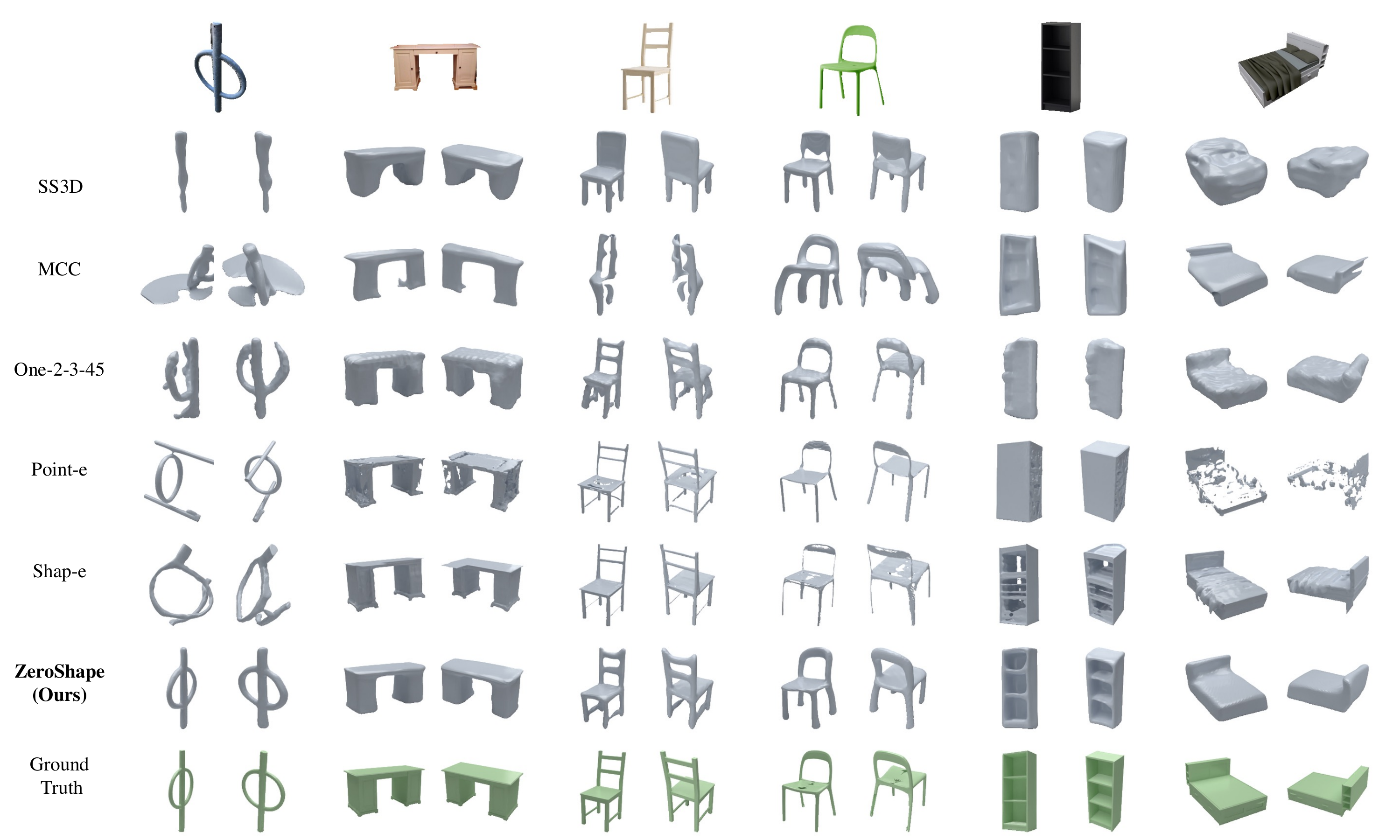} 
    \caption{Qualitative results and comparison on Pix3D.}
    \label{fig:quali-pix3d-supp}
\end{figure*}

\begin{figure*}[h]
\centering
    \includegraphics[width=\linewidth]{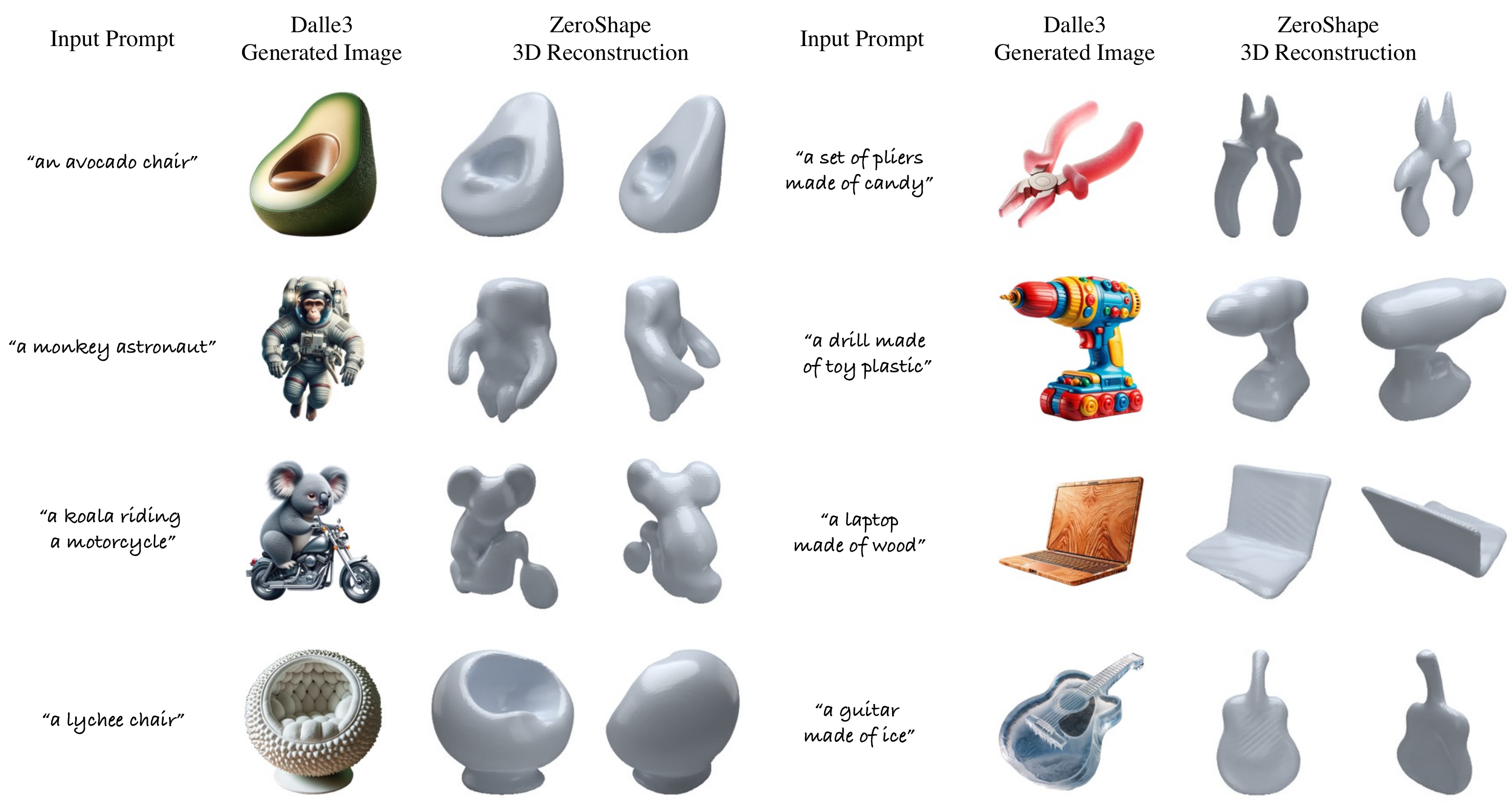} 
    \caption{Qualitative results on images generated with DALL$\cdot$E 3. These results demonstrate the zero-shot generalization ability of ZeroShape to complex novel images.}
    \label{fig:quali-dalle3-supp}
\end{figure*}

\section{Additional Qualitative Comparison}
\label{sec:addn_comparison}
We show additional qualitative results on OmniObject3D, Ocrtoc3D and Pix3D in~\cref{fig:quali-omni-supp},~\cref{fig:quali-ocrtoc-supp} and~\cref{fig:quali-pix3d-supp}, respectively.
Comparing with prior arts, the reconstruction of ZeroShape better captures the global shape structure and visible geometric details.

\section{Inference on AI-generated Images}
\label{sec:generated_inference}
We present additional results of ZeroShape using images generated with DALL$\cdot$E 3. To test the out-of-domain generalization ability, we generate images of imaginary objects as the input to our model (see~\cref{fig:quali-dalle3-supp}). Despite the domain gap to realistic or rendered images, ZeroShape can faithfully recover the global shape structure and accurately follow the local geometry cues from the input image. These results also demonstrate the potential of using ZeroShape in a text-based 3D generation workflow.

\section{Data Curation Details}
\label{sec:data_details}
In this section we describe our data generation procedure for training and for rendering the object scans from OmniObject3D to generate one of our benchmark test sets.
\subsection{Synthetic Training Dataset Generation}
\begin{figure*}[ht]
\centering
    \includegraphics[width=\linewidth]{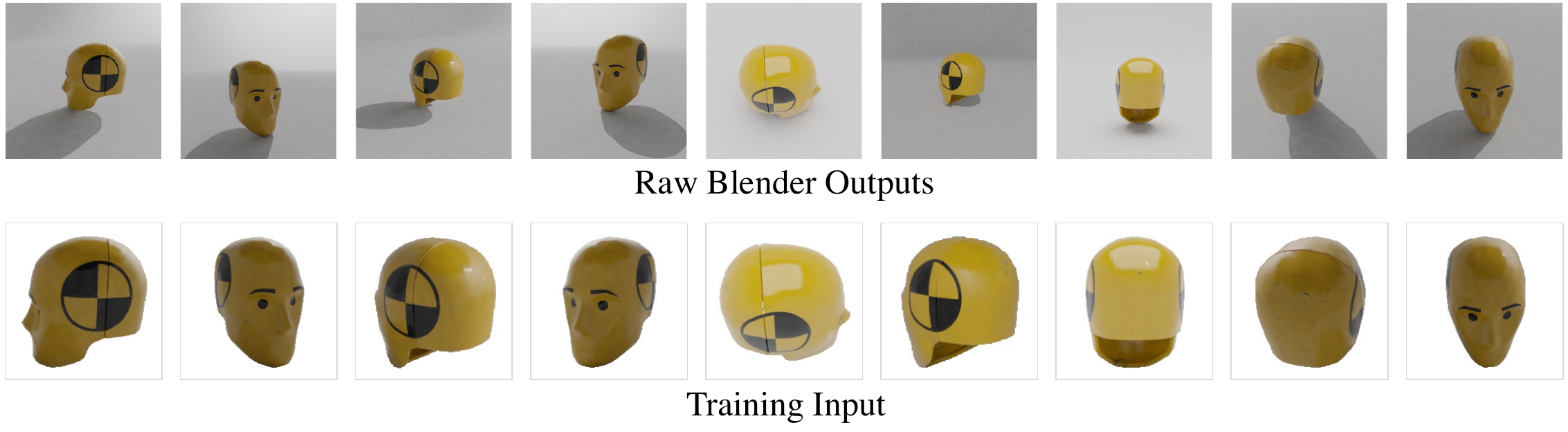} 
    \caption{\textbf{Synthetic Training Data Generation.} We render training images with varying lighting, camera intrinsics and extrinsics. The images are center-cropped, foreground-segmented and resized before being used as training input.}
    \label{fig:training-data-generation}
\end{figure*}

\noindent\textbf{Image Rendering.} 
For an arbitrary 3D mesh asset, our Blender-based rendering pipeline first loads it into a scene and normalizes it to fit inside a unit cube. Our scene consists of a large rectangular bowl with a flat bottom, a common scene setup that 3D artists use for rendering to allow for realistic shading, and 4 point light sources and one area light source. We randomly place cameras around the object with 30mm to 70mm focal length for a 35mm sensor size equivalent. We randomly vary the distance, elevation (from 5 to 65 degrees), the LookAt point of the camera and generate images of $600\times600$ resolution (see~\cref{fig:training-data-generation}). This variation in object/camera geometry allows capturing the variability of projective geometry in real world scenarios, coming from different capture devices and camera poses. This is in contrast with prior work that uses fixed intrinsics, fixed distance, and LookAt pointed at the center of the object. 

In addition to RGB images, we extract segmentation masks, depth maps, intrinsics, extrinsics and object pose. We center crop the objects, mask out the background, resize images to $224\times224$ and process the additional annotations to account for the crop, segmentation and resize.

\noindent\textbf{Ground Truth Occupancy Extraction.} 
To obtain ground truth occupancy, we first extract watertight meshes using the code from occupancy networks~\footnote{\url{https://github.com/autonomousvision/occupancy_networks}}, and then extract SDF for $32^3$ query points per mesh following DISN~\footnote{\url{https://github.com/laughtervv/DISN}}. The SDF is converted to occupancy during training. 

\subsection{Generating the OmniObject3D Testing Set} 
\begin{figure*}[ht]
\centering
    \includegraphics[width=\linewidth]{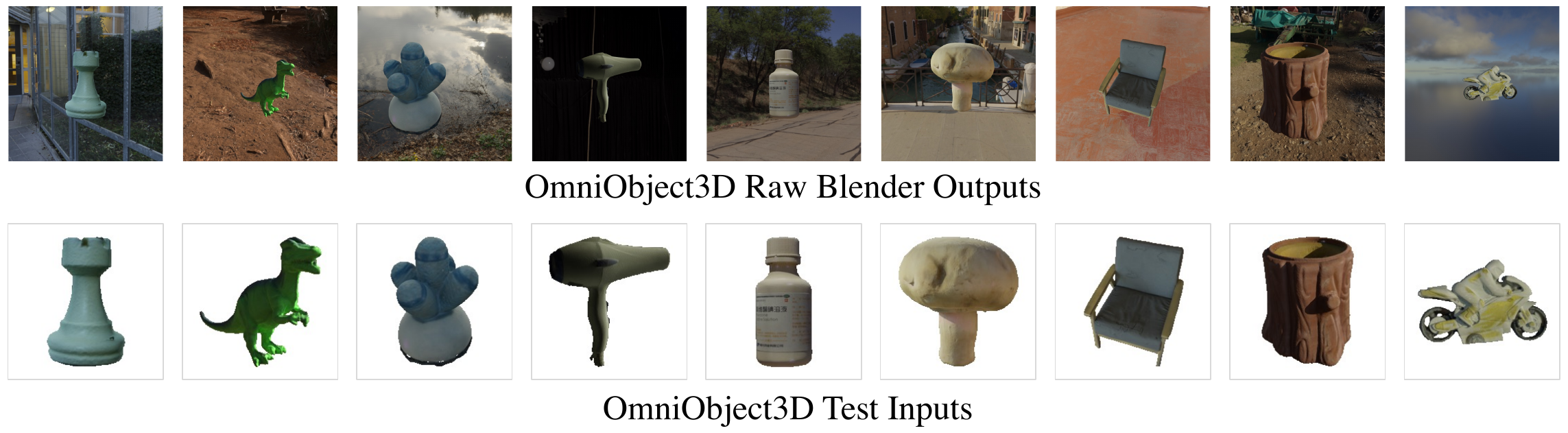} 
    \caption{\textbf{OmniObject3D Testing Data Generation.} For OmniObject3D, we generate realistic testing images with varying lighting, camera intrinsics and extrinsics. To increase rendering realism and diversity, we use diverse HDRI environment maps for scene lighting.}
    \label{fig:testing-data-generation}
\end{figure*}

The original videos released by the OmniObject3D dataset have noisy foreground masks and are mostly taken indoor on a tabletop. To improve the lighting variability and ensure accurate segmentations, we follow the rendering procedure described in the previous section to generate testing data. Different from our training set generation, we use HDRI environment maps to generate scene lighting, which results in high lighting quality and diversity (see~\cref{fig:testing-data-generation}). 

\end{document}